# DensiThAI, A Multi-View Deep learning Framework for Breast Density Estimation using Infrared Images

Siva Teja Kakileti*, Geetha Manjunath

Niramai Health Analytix Pvt Ltd, Bangalore, India, sivateja@niramai.com

**Abstract**

Breast tissue density is a key biomarker of breast cancer risk and a major factor affecting mammographic sensitivity. However, density assessment currently relies almost exclusively on X-ray mammography, an ionizing imaging modality. This study investigates the feasibility of estimating breast density using artificial intelligence over infrared thermal images, offering a non-ionizing imaging approach . The underlying hypothesis is that fibroglandular and adipose tissues exhibit distinct thermophysical and physiological properties, leading to subtle but spatially coherent temperature variations on the breast surface. In this paper, we propose DensiThAI, a multi-view deep learning framework for breast density classification from thermal images. The framework was evaluated on a multi-center dataset of 3,500 women using mammography-derived density labels as reference. Using five standard thermal views, DensiThAI achieved a mean AUROC of 0.73 across 10 random splits, with statistically significant separation between density classes across all splits ($p \ll 0.05$). Consistent performance across age cohorts supports the potential of thermal imaging as a non-ionizing approach for breast density assessment with implications for improved patient experience and workflow optimization.

1. Introduction

Breast density describes the relative proportion and spatial distribution of fibroglandular tissues and adipose tissues within the breast and is recognized as a clinically significant biomarker of breast cancer risk. Epidemiological studies consistently report that women with high breast density exhibit approximately a two- to three-fold increased risk of developing breast cancer compared to those with predominantly fatty breasts [1–4]. Beyond its role in risk stratification, breast density poses a fundamental challenge to conventional X-ray mammography. Dense fibroglandular tissue attenuates X-rays in a manner similar to malignancies, thereby obscuring lesions and substantially reducing sensitivity. Consequently, women with heterogeneously dense or extremely dense breasts experience reduced screening performance and a higher incidence of interval cancers [5]. Reflecting its clinical importance, recent regulatory mandates require explicit communication of breast density information following mammographic screening, underscoring the growing need for density assessment [6].

Current clinical assessment of breast density remains almost exclusively dependent on mammography. In practice, women with dense breasts often require adjunct imaging modalities such as ultrasound or magnetic resonance imaging to address the reduced sensitivity or inconclusive findings of mammography in this population [7, 8]. Consequently, mammography continues to be performed as part of routine screening even in women with dense breasts, where tissue-related masking may limit diagnostic yield. This reliance on an ionizing modality, particularly in scenarios where sensitivity is intrinsically reduced, motivates the investigation of non-ionizing approaches capable of estimating breast tissue composition through alternative physical mechanisms.

Infrared thermal imaging provides a non-contact means of measuring skin-surface temperature distributions with high spatial and thermal resolution. Advances in long-wave infrared sensor technology, including improved noise-equivalent temperature difference and radiometric stability, have renewed interest in thermography as a quantitative imaging modality rather than a purely qualitative screening tool [9]. Unlike structural imaging, infrared thermography reflects the integrated effects of subsurface heat generation, perfusion-mediated heat exchange, and conductive heat transfer through heterogeneous tissue layers.

While breast thermography has historically been explored primarily for the detection of focal abnormalities associated with malignancy [10,11], comparatively little attention has been paid to its potential sensitivity to diffuse tissue composition such as breast density. Fibroglandular and adipose tissues differ markedly in their thermophysical and physiological properties, suggesting that variations in their relative abundance may imprint characteristic, spatially coherent thermal patterns at the surface. Although the resulting temperature differences are expected to be small and influenced by individual anatomy, geometry, and boundary conditions, the governing physics provides a rationale for their detectability.

In this study, we investigate the feasibility of estimating breast density from thermal infrared images. Section 2 outlines the thermophysical mechanisms linking tissue composition to surface temperature distributions for establishing theoretical motivation. Building on this foundation, Section 3 presents a data-driven DensiThAI methodology designed to exploit subtle density-dependent thermal signatures using deep learning.

## 2. Thermophysical Basis for Density-Dependent Thermal Signatures

Breast tissue is a heterogeneous medium composed primarily of adipose and fibroglandular structures, each characterized by distinct thermophysical and physiological properties. These differences influence heat generation, conduction, and removal within the breast, thereby modulating the steady-state temperature distribution observed at the skin surface. This section outlines the physical basis for density-dependent surface temperature signatures.

## 2.1 Governing Bioheat Model

Let the breast domain $\Omega \subset R^3$ comprise adipose and fibroglandular tissues with spatially varying proportions. Under steady-state conditions, the temperature field $T(x)$ satisfies the Pennes bioheat equation:

$$\nabla \cdot (k(x)\nabla T(x)) + \rho_b c_b \omega(x)(T_b - T(x)) + Q(x) = 0, x \in \Omega. \quad (1)$$

Here, $k(x)$ denotes the effective thermal conductivity of tissue; $\omega(x)$ is the volumetric blood perfusion rate; $Q(x)$ represents metabolic heat generation; and $T_b$ is the arterial blood temperature. The parameters $\rho_b$ and $c_b$ are the density and specific heat capacity of blood, respectively. Together, these terms model the combined effects of conductive heat transfer, perfusion-mediated heat exchange, and internal metabolic heat production that govern subsurface-to-surface thermal propagation.

Let $f(x) \in [0,1]$ denote the local fibroglandular fraction. The effective thermophysical properties are approximated by standard linear mixing:

$$k(x) = fk_g + (1-f)k_a, \omega(x) = f\omega_g + (1-f)\omega_a, Q(x) = fQ_g + (1-f)Q_a. \quad (2)$$

Where the subscripts *g* and *a* refer to fibroglandular and adipose tissues, respectively. This approximation provides a first-order representation of spatially varying tissue composition without imposing idealized geometric assumptions.

## 2.2 Thermophysical Contrasts Between Adipose and Glandular Tissues

Equations (1)–(2) indicate that the surface temperature field is governed primarily by tissue thermal conductivity, blood perfusion, metabolic heat generation, volumetric heat capacity, and density. These parameters differ substantially between adipose and fibroglandular tissues, as documented across multiple experimental and modeling studies [12–14]. Table I summarizes representative contrasts in these thermophysical properties. In general, glandular tissue exhibits higher thermal conductivity and metabolic activity, whereas adipose tissue is more insulating and less metabolically active. Consequently, the spatial arrangement, thickness, and relative proportion of these tissue constituents modulate subsurface thermal gradients that ultimately propagate toward the skin surface.

*Table I: Representative thermophysical properties of adipose and fibroglandular breast tissues.*

| | Thermal conductivity $k$ (W/m·K) | Metabolic heat $Q_m$ (W/m³) | Specific heat capacity c (J/Kg/K) | Blood perfusion rate $\omega_t$ (kg·s⁻¹·m⁻³) | Density (kg/m³) |
|---|---|---|---|---|---|
| adipose | 0.171 | 458 | 2220 | 0.886 | 932 |
| glandular | 0.328 | 1180 | 3398 | 0.886 | 1066 |

Multiple computational studies have demonstrated that adipose–glandular heterogeneity can induce small but measurable surface temperature variations. For example, González et al. [12] and Asnida et al. [14] reported surface temperature differences on the order of 0.15–0.5 °C arising from variations in internal tissue composition under controlled conditions.

Prior work has shown that deep-learning models can extract subtle thermal patterns linked to underlying physiological variations, including malignancy-associated changes, even when absolute temperature differences are extremely small [10–11,15–16]. In bioheat modeling, malignant tumors are often approximated as localized glandular inclusions with elevated metabolic heat generation and perfusion, leading to detectable surface perturbations. While breast density might not involve focal extremes or localized angiogenic heating, the systematic thermophysical contrasts between adipose and fibroglandular tissues suggest that density-related information may nonetheless be encoded in the surface temperature field.

Accordingly, under the bioheat framework described above, we hypothesize that high-resolution infrared thermography, when coupled with artificial intelligence, may encode information related to breast tissue density through thermally mediated contrasts between adipose and fibroglandular tissues. While deep learning models do not explicitly implement the bioheat equation, they operate on surface temperature distributions shaped by the same diffusive and perfusion-mediated mechanisms governing subsurface heat transport.

### 3. Proposed Methodology

Building on the thermophysical observations described in Section 2, our methodology aims to determine whether high-resolution infrared thermography contains sufficient information to infer underlying breast tissue composition. Typically, breast thermography involves capturing thermal images at five standard views comprising frontal (0°), left lateral (90°), right lateral (-90°), left oblique(45°) and right oblique views (-45°). These complementary views capture 3D surface spatial variations in temperature distribution and allows to study geometrical and depth related thermal variations.

We propose DensiThAI, a multi-view learning framework for inferring breast tissue density from infrared thermography. DensiThAI is designed to capture density-dependent thermal organization that arises from differences in metabolic heat generation, perfusion, and thermal diffusion. The framework consists of four components: view normalization, multi-view feature extraction, multi-view fusion, and supervised density learning as illustrated in Figure 1.

### 3.1 View Normalization

Let $T_v(x)$ denote the radiometric temperature maps for view $v \in \{1, ..., V\}$. Since density-related thermal effects are expected to manifest as relative spatial gradients rather than absolute temperature offsets, all views of a given subject are normalized using a common subject-specific temperature range. Specifically, normalization is performed with respect to the minimum and maximum temperature observed across all five views of the subject.

$$\widetilde{T}_v(x) = \frac{T_v(x) - T^s_{min}}{T^s_{max} - T^s_{min}} \qquad (3)$$

$T^s_{max}$ and $T^s_{min}$ are subject level minimum and maximum temperatures. This normalization preserves cross-view thermal consistency while reducing inter-subject variability.

### 3.2 Multi-View Feature Encoding

Each normalized view $\widetilde{T}_v$ is then processed using a convolutional encoder based on the VGG-16 architecture [17]. All input views are normalized to 224x224 and replicated across three channels to form 224x224x3 inputs. The top classification layer is removed and the output of the convolution block is fed to the global average pooling layer, yielding a 512 dimensional latent representation.

$$z_v = f_\theta\left(\widetilde{T}_v\right) \in R^{512} \qquad (4)$$

Weight sharing across views enforces view-invariant feature learning, reflecting the assumption that tissue composition induces consistent thermophysical effects across viewing angles. All weights are initialized from ImageNet pretraining and fine-tuned on thermal data.

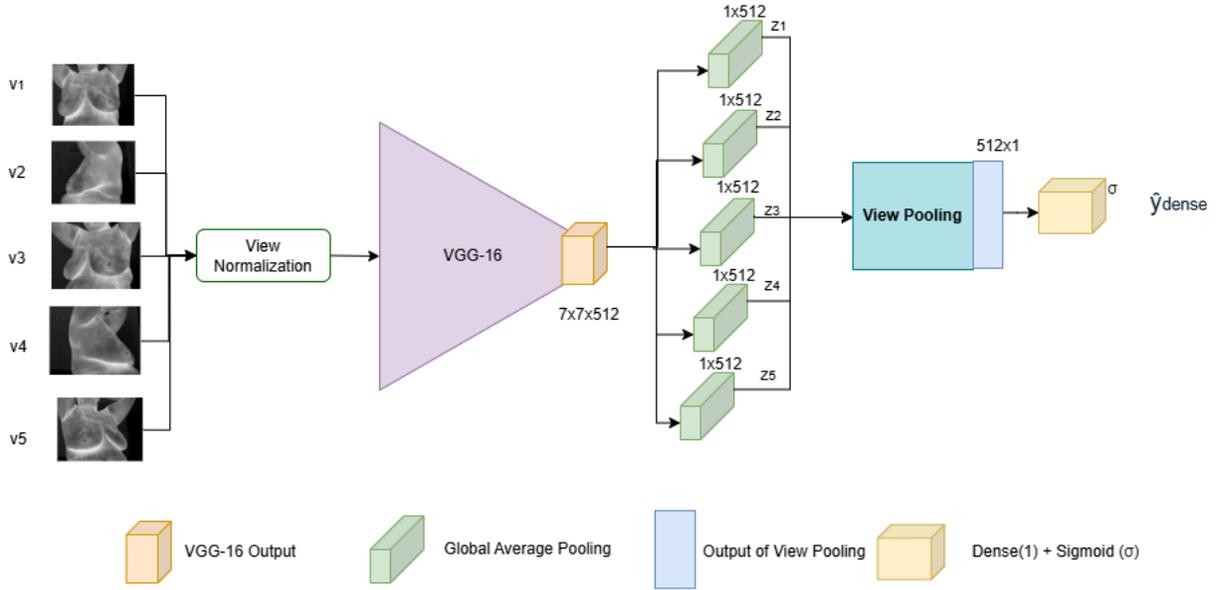

**Figure 1:** Block diagram illustrating the multi-view DensiThAI framework employed for density estimation.

### 3.4 Multi-View Pooling

To integrate complementary information across the multiple thermal views, DensiThAI employs a simple average pooling strategy across views. Specifically, the latent representations extracted from each view are aggregated by computing their mean.

$$z = \frac{1}{V} \sum_{v=1}^{V} z_v \qquad (5)$$

This pooling operation treats all views equally and provides a stable fused representation that summarizes global thermal patterns across different viewing angles. Average pooling was chosen to avoid introducing additional trainable parameters and to reduce the risk of overfitting, while still preserving density-related information distributed across views.

### 3.5 Density Classification

The fused representation $z$ is mapped to a probabilistic estimate of breast density via a linear projection followed by a sigmoid activation:

$$\hat{y} = \sigma(w^\top z + b) \tag{6}$$

where $\hat{y} \in (0, 1)$ denotes the predicted probability of dense breast tissue. Model parameters are optimized using the binary cross-entropy loss,

$$L_{BCE} = -[y \log(\hat{y}) + (1 - y)\log(1 - \hat{y})] \tag{7}$$

with $y \in \{0, 1\}$ the ground-truth label.

## 4. Results

### 4.1 Dataset Description

The proposed DensiThAI framework was developed and evaluated using a multi-center clinical dataset comprising 3,500 women with breast density annotations available from routine mammography reports. Data collection was conducted across multiple hospitals following approval from the respective Institutional Ethics Committees, and written informed consent was obtained from all participants prior to imaging.

Women undergoing mammographic screening were offered an additional infrared thermal imaging test on a voluntary basis. Thermal image acquisition followed a standardized protocol involving five canonical views: frontal (0°), left lateral (90°), right lateral (−90°), left oblique (45°), and right oblique (−45°). Subject preparation, environmental conditions, and imaging procedures were controlled in accordance with the guidelines recommended by the American Academy of Thermology [18], ensuring inter-center consistency and minimizing non-physiological sources of thermal variability. Thermal images were acquired using FLIR infrared cameras with a spatial resolution of 240 × 320 pixels and a thermal sensitivity lower than 50 mK.

Breast density labels were obtained from the corresponding mammography reports. Density reporting practices varied across institutions: some adhered to the ACR BI-RADS density categories (A–D), while others used descriptive terminology (predominantly fatty, scattered fibroglandular, heterogeneously dense, extremely dense). For analytical consistency, these annotations were harmonized into a binary density classification. Specifically, ACR A/B or predominantly fatty/scattered fibroglandular breasts were classified as fatty, while ACR C/D or heterogeneously dense/extremely dense breasts were classified as dense. Sample thermal images of both categories were illustrated in Figure 2.

Using this harmonization, 2,435 subjects were labeled as fatty and 1,065 subjects as dense. To assess the feasibility and generalizability of breast density estimation from thermal infrared images, the dataset was partitioned at the subject level into training (60%), validation (20%), and independent test (20%) subsets. To mitigate bias associated with any single random split, this partitioning was repeated across 10 independent random seeds, and all reported results correspond to the mean ± standard deviation across these splits. A summary of the dataset composition is provided in Table I, stratified by age group.

|  | Total Subjects | Class: Fatty | Class: Dense |
|---|---|---|---|
| Overall | 3,500 | 2,435 | 1,065 |
| Age <=45 | 1,029 | 530 | 499 |
| Age>45 | 2,471 | 1,905 | 566 |

**Table I:** Dataset distribution across population and age cohorts <=45 and >45 years

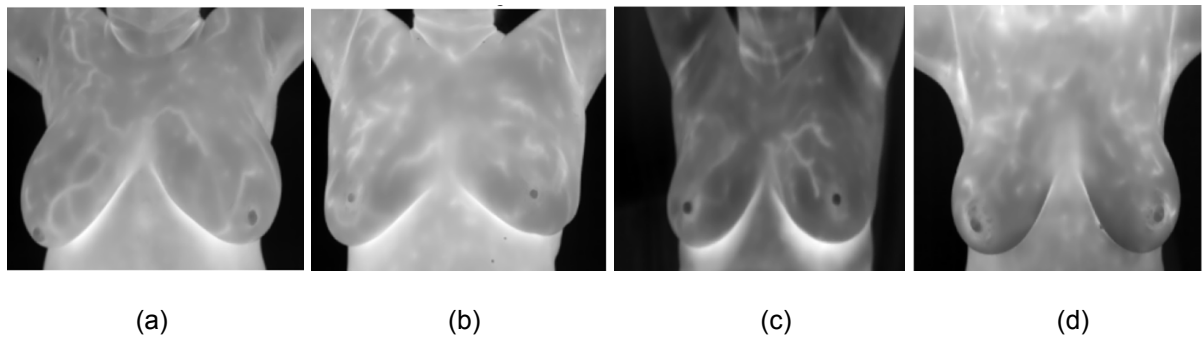

(a)        (b)        (c)        (d)

**Figure 2:** Sample breast thermal images of women classified as (a-b) fatty breast tissue (c-d) dense breast tissue.

### 4.2 Overall Classification Performance

To evaluate the feasibility of breast density discrimination from thermal infrared images, we applied the proposed DensiThAI framework using two primary input configurations:
 (i) full-field thermal images with multi-view input, and
 (ii) cropped breast-region thermal images with multi-view input.

In the full-field configuration, all five thermal views were used directly as input without anatomical cropping. This configuration allows the model to utilize thermal cues not only from the breast region but also from adjacent anatomical areas such as the axilla, neck, and upper abdomen. In the cropped configuration, breast bounding boxes were extracted using segmentation masks generated by a cascaded convolutional neural network [19], followed by manual refinement by trained technicians to ensure anatomical accuracy. Each bounding box comprising the breast region was treated as an independent sample; consequently, frontal views yielded separate left and right breast regions, resulting in six breast regions per subject. This preprocessing restricts the thermal information strictly to the breast area.

As summarized in Table II, both input configurations demonstrated statistically significant discrimination between fatty and dense breast classes. Using full-field thermal images, DensiThAI achieved an area under the receiver operating characteristic curve (AUROC) of $0.73 \pm 0.016$ and an area under the precision–recall curve (AUPRC) of $0.56 \pm 0.023$. Comparable performance was observed when using cropped breast-region images, with AUROC and AUPRC values of $0.72 \pm 0.026$ and $0.53 \pm 0.040$, respectively.

To assess the statistical significance of the observed separation between classes, a Mann–Whitney U test was performed on the predicted density scores. The resulting p-values ($p < 1 \times 10^{-17}$ for full-field images and $p < 1 \times 10^{-11}$ for cropped images) confirm statistically significant separation between fatty and dense breasts. These results indicate that thermal infrared images contain sufficient information to enable consistent and statistically significant discrimination of breast density.

In addition to the multi-view learning framework, we also trained single-view DensiThAI models, where each thermal view was processed independently using the same network architecture. For subject-level density estimation, predictions from individual views were aggregated using a maximum operation across all available views. The resulting performance metrics are reported in Table II.

For comparison, a baseline model was developed using handcrafted radiomics features extracted with the PyRadiomics library (v3.0.1) [20]. First-order statistical and gray-level co-occurrence matrix (GLCM) texture features were extracted from each view and concatenated across views to form a 210-dimensional feature vector per subject. Random forest classifiers trained on these features exhibited substantially lower discrimination performance compared to DensiThAI, as reported in Table II, highlighting the advantage of proposed multi-view thermal feature learning for density estimation.

### 4.3 Age-cohort Classification Performance

An age-cohort analysis was conducted to evaluate whether age-related anatomical characteristics, such as changes in breast contour, body contour and sagging, introduce confounding cues that could influence breast density classification performance. To evaluate this, model performance was assessed separately for two age groups: ≤45 years and ≥45 years.

As summarized in Table II, DensiThAI demonstrated comparable discrimination performance across both age cohorts for all evaluated input configurations. The AUROC and AUPRC values were broadly consistent between both age cohorts participants, indicating that the learned thermal representations are not strongly driven by age-dependent anatomical variations. Furthermore, statistically significant separation between fatty and dense classes was preserved within each age group, as reflected by the Mann–Whitney U test results reported in Table II.

These findings suggest that the proposed framework generalizes well across age cohorts and that breast density discrimination using thermal infrared imaging is not substantially confounded by age-related morphological differences.

|  |  | AUROC | AUPRC | Mann–Whitney U test |
|---|---|---|---|---|
| Multi-view DensiThAI with full-field thermal image as input | Overall | 0.73±0.016 | 0.56±0.023 | $p<1 \times 10^{-17}$ |
|  | Age<=45 | 0.72±0.025 | 0.70±0.035 | $p<1 \times 10^{-5}$ |
|  | Age>45 | 0.70±0.022 | 0.43±0.032 | $p<1 \times 10^{-7}$ |
| Multi-view DensiThAI with cropped thermal image as input | Overall | 0.72±0.026 | 0.53±0.040 | $p<1 \times 10^{-11}$ |
|  | Age<=45 | 0.68±0.027 | 0.66±0.054 | $p<1 \times 10^{-4}$ |
|  | Age>45 | 0.70±0.033 | 0.44±0.050 | $p<1 \times 10^{-5}$ |
| Single-view DensiThAI with full-field thermal image as input | Overall | 0.73±0.014 | 0.55±0.024 | $p<1 \times 10^{-16}$ |
|  | Age<=45 | 0.71±0.033 | 0.70±0.043 | $p<1 \times 10^{-4}$ |
|  | Age>45 | 0.70±0.021 | 0.43±0.036 | $p<1 \times 10^{-7}$ |
| Single-view DensiThAI with cropped thermal image as input | Overall | 0.72±0.016 | 0.54±0.020 | $p<1 \times 10^{-15}$ |
|  | Age<=45 | 0.69±0.024 | 0.67±0.037 | $p<1 \times 10^{-3}$ |
|  | Age>45 | 0.71±0.017 | 0.43±0.031 | $p<1 \times 10^{-8}$ |
| Radiomics based Density Estimation with full-field multi-view thermal images | Overall | 0.62±0.022 | 0.43±0.026 | $p<1 \times 10^{-11}$ |
|  | Age<=45 | 0.63±0.043 | 0.60±0.052 | $p<1 \times 10^{-5}$ |
|  | Age>45 | 0.60±0.027 | 0.34±0.036 | $p<1 \times 10^{-5}$ |

**Table II:** Performance of the proposed DensiThAI framework for breast density classification using full-field and anatomically cropped thermal images under single-view and multi-view deep learning

configurations, compared with a radiomics-based baseline. An AUROC of 0.5 corresponds to random classification, while AUPRC values near the class prevalence (0.30 overall, 0.48 for age ≤45 years, and 0.23 for age >45 years) indicate chance-level performance. Mann–Whitney U test with $p < 0.05$ considered statistically significant.

## 4.4 Temporal Stability of Density Scores

To assess the short-term consistency of the proposed DensiThAI framework, we performed a repeat-visit analysis on an independent cohort of 1,512 participants, each of whom had two thermal imaging sessions acquired within 30 days. This cohort was distinct from the labeled density dataset and did not include radiologist-provided breast density annotations.

We evaluated the multi-view DensiThAI model using full-field thermal images and quantified the absolute difference in predicted density scores between the two visits for each participant. Across subjects, the mean absolute inter-visit score difference was 0.03, with a variance of 0.001, indicating high short-term stability of the predicted density scores. The observed stability suggests that the proposed framework produces reproducible density estimates over short time intervals, supporting its robustness to acquisition variability and day-to-day physiological fluctuations.

## 5. Discussion

Breast density is a clinically important biomarker that influences both breast cancer risk and the performance of mammographic screening. This study demonstrates, for the first time to our knowledge, that breast tissue density can be discriminated using thermal infrared imaging alone, without reliance on ionizing radiation or structural imaging. By combining high resolution thermal imaging with a multi-view deep learning DensiThAI framework, we show that subtle, spatially distributed thermal patterns encode information related to underlying breast tissue composition.

Fibroglandular and adipose tissues differ in thermal conductivity, metabolic heat generation, perfusion, and heat capacity, leading to systematic differences in subsurface heat transport. Unlike malignancies, breast density does not produce localized thermal extremes; rather, it represents a diffuse and spatially heterogeneous alteration in tissue composition. Consequently, density-related information is not expected to manifest as absolute temperature elevations but instead as subtle, spatially coherent surface patterns and gradients (illustrated in Figure 2). The ability of the proposed framework to detect such patterns suggests that physiologically meaningful thermal organization persists despite inter-subject variability in anatomy, boundary conditions, and environmental influences.

A key aspect of the proposed approach is the use of standardized multi-view thermal imaging. Different viewing angles sample complementary surface regions and implicitly encode depth-weighted information influenced by breast geometry, tissue thickness, and internal tissue composition. The comparable performance obtained using full-field and anatomically cropped breast-region inputs suggests that density-related information is primarily localized to the breast itself, while broader contextual cues from surrounding anatomy may provide secondary complementary information. Furthermore, the marginal improvement observed over single-view configurations supports the importance of integrating thermal information across multiple views for robust breast density discrimination.

Potential confounding factors were further evaluated through age-stratified analysis. The consistent performance observed across age cohorts (≤45 years and ≥45 years) suggests that the model does not rely predominantly on age-related anatomical features such as breast sagging or external contour changes. Together with the cropped-input experiments, which reduce the influence of body shape and non-breast anatomy, these findings indicate that the proposed framework primarily captures intrinsic thermal patterns associated with tissue composition rather than superficial, demographic, or geometric correlates.

The proposed deep learning framework substantially outperformed conventional radiomics-based approaches relying on handcrafted texture features. This observation aligns with broader experience in medical imaging, where deep convolutional models have proven more effective at capturing complex, multiscale spatial dependencies. In the context of breast density estimation, such higher-order spatial organization and cross-view consistency appear essential, as density-related thermal effects are distributed and subtle rather than localized and high-contrast.

Beyond technical performance, these findings suggest a potential role for thermal AI–based density estimation within future breast screening workflows. The predicted density score could, in principle, be incorporated into established risk-prediction frameworks (e.g., Tyrer–Cuzick [21]) to support density-aware screening strategies and inform the selection of adjunct imaging modalities. Recent studies have further suggested that multi-modal screening approaches such as mammography for predominantly fatty breasts and AI-assisted thermographic analysis for dense breasts may improve overall screening performance relative to either modality alone [22]. Figure 3 illustrates a conceptual workflow in which thermal density estimation acts as a non-ionizing triage and stratification layer within multimodal screening pathways. While such integration remains exploratory and requires prospective validation, the present results provide a quantitative and physics-consistent foundation for considering thermal infrared imaging as a non-ionizing adjunct for breast density assessment.

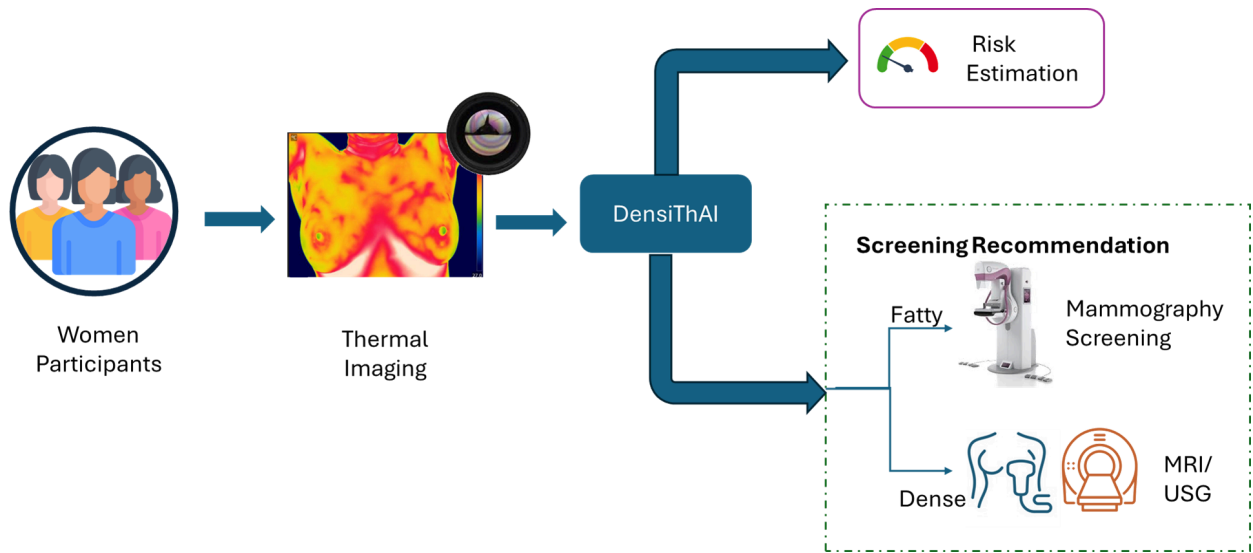

**Figure 3.** Potential operational workflows of utilizing DensiThAI density in the real-world.

Several limitations merit consideration. Breast density labels were derived from routine clinical mammography reports, which are known to exhibit inter-reader variability and institution-dependent reporting practices. As a result, the training labels likely contain an inherent degree of subjectivity and noise. The ability to achieve statistically significant discrimination despite this variability suggests that the model is capturing robust, physiologically meaningful thermal patterns rather than overfitting to idiosyncratic labeling. Future studies would benefit from training against quantitative or volumetric mammographic density measures derived directly from imaging data, which may enable finer-grained analysis and improved performance. In the present study, such data were not available, as only report-level density annotations were accessible. Further, large-scale validation needs to be performed to evaluate the true clinical performance of density stratification with thermal imaging.

## Conclusion

This study establishes that breast density, despite being a diffuse and non-localized tissue property, leaves a detectable thermal signature at the skin surface that can be captured through multi-view infrared imaging and learned using data-driven methods. By integrating AI with infrared thermal imaging, the proposed framework demonstrates that non-ionizing thermal measurements contain information relevant to radiologically defined breast density, even in the presence of inter-subject variability and noisy clinical labels. These findings broaden the conceptual scope of breast thermography from focal abnormality detection to global tissue characterization and motivate further investigation using quantitatively defined density measures and physiologically informed modeling. Future work should focus on large scale validation, multi-class classification and also quantified mammography density reporting.